\PassOptionsToClass{balance=false}{acmart}
\documentclass[sigconf,nonacm]{acmart}

\AtBeginDocument{}
\setcopyright{none}
\acmDOI{}
\acmISBN{}
\usepackage{booktabs}
\usepackage{placeins}
\usepackage{tabularx,colortbl}
\usepackage{pifont}            % \ding{172}.. circled numbers, matching the figure
\usepackage{xspace}
\usepackage{algorithm}
\usepackage{algpseudocode}

\newcommand{\method}{MHVF\xspace}

\begin{document}
\raggedbottom

\title[MHVF: Training-Free GUI Grounding]{MHVF: Training-Free GUI Grounding via Verification-Guided Multi-Hypothesis Search}

% arXiv author information. The anonymous conference manuscript is separate.
\newcommand{\MHVFNameA}{Minghua Chen}
\newcommand{\MHVFNameB}{Xingming Liao}
\newcommand{\MHVFNameC}{Weiqiang Zhong}
\newcommand{\MHVFNameD}{Zhuowei Wang}
\newcommand{\MHVFEmailA}{2232706035@mails.gdut.edu.cn}
\newcommand{\MHVFEmailB}{liaoxingming@mails.gdut.edu.cn}
\newcommand{\MHVFEmailC}{zhongweiqiang@mails.gdut.edu.cn}
\newcommand{\MHVFEmailD}{zwwang@gdut.edu.cn}
\newcommand{\MHVFDepartment}{School of Computer Science and Technology}
\newcommand{\MHVFInstitution}{Guangdong University of Technology}

% Keep one structured author record per person for PDF metadata.
\author{\MHVFNameA}
\email{\MHVFEmailA}
\author{\MHVFNameB}
\email{\MHVFEmailB}
\author{\MHVFNameC}
\email{\MHVFEmailC}
\author{\MHVFNameD}
\authornote{Corresponding author: Zhuowei Wang.}
\email{\MHVFEmailD}
\affiliation{%
  \department{\MHVFDepartment}
  \institution{\MHVFInstitution}
  \city{Guangzhou}
  \country{China}
}
\renewcommand{\shortauthors}{Chen et al.}

% A shared affiliation block for the four authors of this preprint.
% This overrides only the arXiv title block, not the body or the class file.
\makeatletter
\def\@mkauthors@iii{%
  \gdef\@currentauthors{}%
  \gdef\@currentaffiliation{}%
  \global\setbox\mktitle@bx=\vbox{%
    \unvbox\mktitle@bx\par\medskip
    \centering
    {\@authorfont
      \MHVFNameA\quad \MHVFNameB\quad \MHVFNameC\quad
      \MHVFNameD\textsuperscript{*}\par}
    \medskip
    {\@affiliationfont
      \MHVFDepartment\\
      \MHVFInstitution, Guangzhou 510090, China\par}
    \smallskip
    {\normalsize
      \href{mailto:\MHVFEmailA}{\MHVFEmailA}\quad
      \href{mailto:\MHVFEmailB}{\MHVFEmailB}\quad
      \href{mailto:\MHVFEmailC}{\MHVFEmailC}\\
      \href{mailto:\MHVFEmailD}{\MHVFEmailD}\par}
    \bigskip
  }%
}
\makeatother

\begin{abstract}
GUI grounding enables web agents to translate natural-language instructions
into actionable interface locations. However, dense layouts and small,
visually similar controls can lead models toward plausible but incorrect
elements, while refinement around a single prediction may leave alternative
targets unexplored. We propose the Multi-Hypothesis Verification Framework
(\method) for training-free GUI grounding. It combines multi-hypothesis
exploration with local visual verification using a single pretrained
vision-language model. To explore
alternative targets, \method constructs spatial hypotheses from multiple
full-screen predictions and examines the corresponding regions through
zoomed re-localization. To resolve ambiguity among the resulting candidates,
it verifies their agreement with the instruction in marked local views.
Verification scores then guide a second round of local search and rank
candidates retained from both search levels for final selection.
Experiments with five general-purpose and GUI-specific models across eight
benchmark settings demonstrate improvements in grounding accuracy in most
settings. \method improves UI-Venus-Ground-7B by 4.3 and 2.1 percentage points
on Mind2Web and VisualWebBench, respectively, and raises Qwen2.5-VL-7B's
accuracy on ScreenSpot-Pro from 23.6\% to 53.6\%.

\end{abstract}

% CCS identifiers verified against the ACM CCS 2012 taxonomy.


\begin{teaserfigure}
  \centering
  \includegraphics[width=\linewidth]{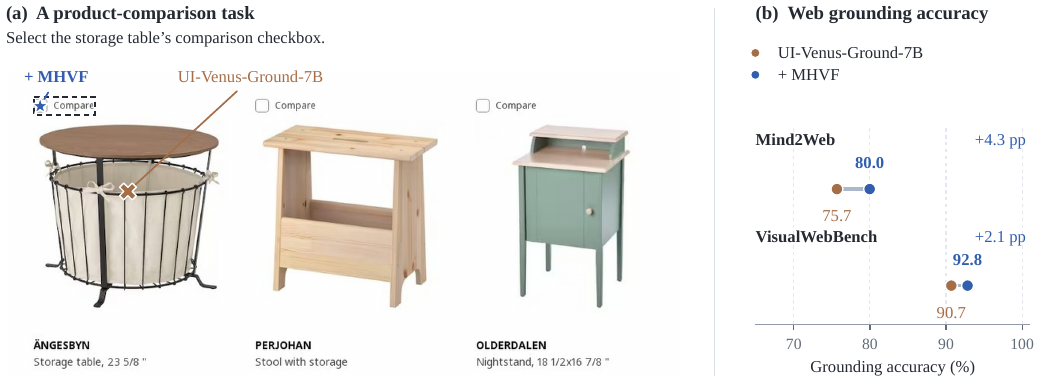}
  \caption{Web grounding with MHVF. (a) On a Mind2Web product-comparison task, UI-Venus-Ground-7B clicks the product image; adding MHVF locates its comparison checkbox. The dashed box marks the target. (b) Grounding accuracy of the same backbone on two web benchmarks; gains are in percentage points.}
  \label{fig:intro_teaser}
  \Description{A product-page example compares the UI-Venus-Ground-7B click on the product image with the MHVF click on its comparison checkbox. Two paired-dot plots show grounding accuracy on Mind2Web and VisualWebBench.}
\end{teaserfigure}

\maketitle

\section{Introduction}
\label{sec:intro}

Web agents rely on GUI grounding to translate natural-language instructions
into clicks on the intended interface elements. Vision-language models
(VLMs) enable this process directly from screenshots, but grounding remains
a major obstacle to reliable web interaction~\cite{zheng2024seeact}.
Dense page layouts, small controls, and visually similar elements make
precise localization challenging, particularly when fine details are lost
at the model's input resolution~\cite{liu2024visualwebbench}.
Figure~\ref{fig:intro_teaser}(a) illustrates such a failure: given an
instruction to select a product's comparison checkbox, the model clicks
the product image.

Recent training-free approaches improve GUI grounding through zooming,
multi-view prediction, and candidate comparison. ZoomClick investigates
staged zooming, MVP aggregates predictions across multiple views, and
AutoFocus combines region proposals with visual candidate
comparison~\cite{jiang2025zoomclick,zhang2025mvp,yao2026autofocus}.
Magnification provides finer visual evidence, but successful grounding
also requires inspecting regions that contain the target and choosing a
candidate that matches the instruction. Initial predictions can guide
this exploration by suggesting several possible target locations, while
verification can assess the candidates obtained through local inspection.
This motivates using local visual verification to connect candidate
assessment with subsequent region search and final selection.

To this end, we propose the Multi-Hypothesis Verification Framework
(\method) for training-free GUI grounding. It combines multi-hypothesis
search with local visual verification using a single pretrained VLM.
At the first level, multiple full-screen
predictions identify potential target regions for inspection. The model
re-localizes the target in magnified views of these regions and verifies
whether each candidate matches the instruction. To focus additional
exploration, \method reopens the region associated with the
highest-scoring candidate for a second round of hypothesis generation,
re-localization, and verification. Candidates from both levels are retained,
and the point with the highest verification score is selected as the final
click.

We conduct extensive experiments with general-purpose and GUI-specific
models across web, desktop, and cross-platform interfaces. The evaluation
covers web grounding on Mind2Web and VisualWebBench, high-resolution
professional interfaces on ScreenSpot-Pro, and cross-platform grounding
on MMBench-GUI L2, among other benchmarks. The results show that \method
improves the grounding accuracy of existing models in most evaluated
settings.

Our main contributions are summarized as follows:
\begin{itemize}
  \item We propose \method, a training-free framework for GUI grounding
  that explores multiple spatial hypotheses and verifies the resulting
  candidates using a single pretrained VLM.
  \item We develop a two-level search mechanism that uses local verification
  scores to guide region reopening and select the final click from
  candidates retained across both levels.
  \item We evaluate \method on five models across eight benchmark settings
  covering web, desktop, and cross-platform interfaces, demonstrating
  improved grounding accuracy in most settings.
\end{itemize}

\section{Related Work}
\label{sec:related}

\subsection{Web Agents and GUI Grounding}
\label{sec:related_grounding}

Web agents use vision-language models to interpret interfaces and translate
user instructions into executable actions. SeeAct investigates this process
by separating action generation from grounding actions to web
elements~\cite{zheng2024seeact}. To support interaction directly from
screenshots, subsequent research develops models that map natural-language
instructions to target locations. SeeClick introduces grounding-oriented
pretraining with automatically collected data, while UGround and OS-Atlas
expand grounding data and model capabilities across web, desktop, and
mobile interfaces~\cite{cheng2024seeclick,gou2025uground,wu2025osatlas}.
These efforts provide the grounding capabilities on which \method builds,
with our focus on improving existing models through inference-time search
and verification.

\subsection{Test-Time Scaling for GUI Grounding}
\label{sec:related_scaling}

Test-time scaling improves grounding by allocating additional computation
to visual inspection and prediction. One line of work explores how to
select and examine informative screenshot regions. ZoomClick studies
staged zooming and the effects of initialization, search depth, and context
retention, while DRS-GUI uses Monte Carlo Tree Search to plan region-level
perceptual actions~\cite{jiang2025zoomclick,liu2026drsgui}. Adaptive methods
further use model predictions to determine when and where to zoom.
UI-Zoomer combines spatial consensus with generation confidence to trigger
zooming and derives crop sizes from prediction
uncertainty~\cite{tang2026uizoomer}. AutoFocus uses visual self-verification
to decide whether refinement is needed, then constructs region proposals
from spatial uncertainty estimated through sampled coordinate
predictions~\cite{yao2026autofocus}.

A related line of work examines how to aggregate or assess multiple
predictions. GUI-RC identifies consensus regions through spatial voting
over sampled predictions, while MVP aggregates predictions from
attention-guided views and selects the centroid of the densest spatial
cluster~\cite{du2026guirc,zhang2025mvp}. Other approaches use visual
reasoning to compare candidates. DiMo-GUI generates candidates through
separate text and icon grounding paths and selects between them using the
instruction and full-resolution screenshot~\cite{wu2025dimogui}.
AutoFocus jointly compares images annotated with candidate locations to
determine the final prediction~\cite{yao2026autofocus}. \method assesses
candidates individually in magnified local views using the same grounding
model. Its verification scores guide both the choice of a region to reopen
for further exploration and the final selection from candidates retained
across two search levels.

% ---------------------------------------------------------------------------
% Method.  Every constant and rule below is checked against
%   10_方法开发_www/ggf/methods.py  (class MHV2, run_mhvf.VARIANTS["mhvf"])
%   10_方法开发_www/ggf/engine.py   (candidates, sigma, score_binary)
% Change the code, change this file.
% ---------------------------------------------------------------------------
\section{Method}
\label{sec:method}

\subsection{Problem Formulation}
\label{sec:formulation}

Given a GUI screenshot $I$ and a natural-language instruction $q$, GUI
grounding aims to predict a click position $p=(x,y)\in[0,1]^2$ on the
interface element specified by $q$. Coordinates are normalized with
respect to the original screenshot.

We use a frozen pretrained vision-language model $G$ as the base grounder.
A view $v$ is either the full screenshot or a local region obtained through
cropping and resizing, with $v_0$ denoting the full-screen view. A
\emph{spatial hypothesis} specifies a region for further inspection. A
\emph{candidate click} is a position obtained by re-localizing the target
within the corresponding local view and mapping the prediction back to
the original screenshot.

\subsection{Overview of MHVF}
\label{sec:overview}

\begin{figure*}[t]
  \centering
  \includegraphics[width=\linewidth]{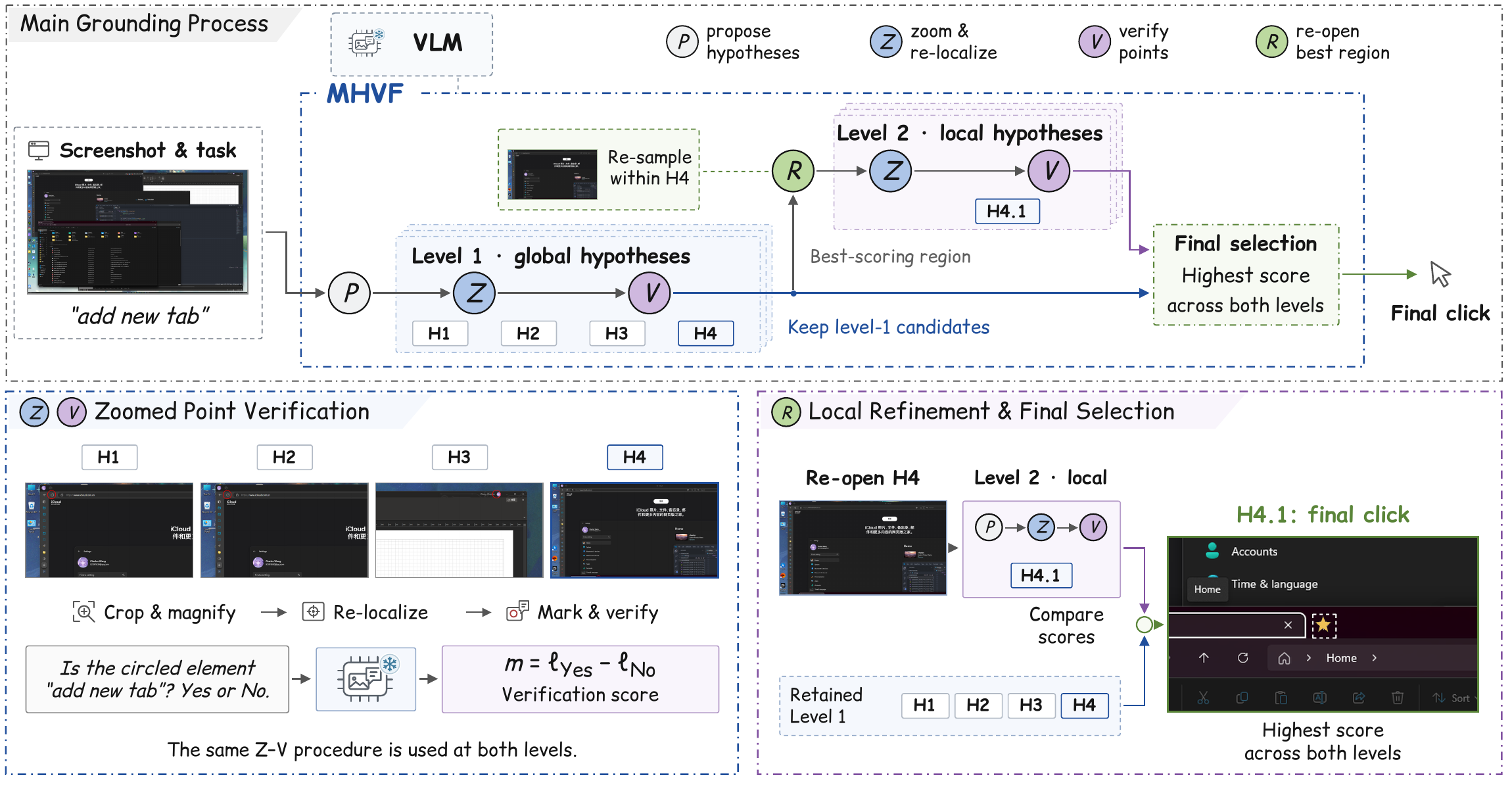}
  \caption{Overview of \method. The blue boundary encloses the inference
  procedure, which reuses a shared frozen VLM for hypothesis generation
  (P), zoomed re-localization (Z), and candidate verification (V).
  Reopening (R) refines the best-scoring region, and candidates from both
  levels remain eligible for final selection. The lower panels illustrate
  verification and local refinement for ``add new tab,'' where H4 is
  reopened and H4.1 yields the final click.}
  \label{fig:method}
  \Description{A frozen VLM is shared by the MHVF inference procedure.
  The lower search level proposes and verifies global hypotheses. The upper
  search level refines the best region, while both candidate sets remain
  available for final selection. Two lower panels illustrate the local
  verification and refinement steps.}
\end{figure*}

As illustrated in Figure~\ref{fig:method}, \method organizes grounding into
two search levels. At level~1, the model produces multiple full-screen
predictions, which are grouped into spatial hypotheses (\emph{P}). Each
hypothesis determines a region to crop and magnify, allowing the model to
re-localize the target with more detailed visual evidence (\emph{Z}). The
resulting candidate is then marked in its local view and assessed against
the instruction (\emph{V}).

At level~2, \method reopens the view associated with the highest-scoring
first-level candidate (\emph{R}). The model generates new hypotheses within
this view and applies the same localization and verification procedure to
obtain additional candidates. Candidates from both levels remain available
for final selection, and \method returns the click with the highest
verification score. The same model performs hypothesis generation,
localization, and verification throughout the process.

% ---------------------------------------------------------------------------
\subsection{Spatial Hypothesis Generation}
\label{sec:propose}

A single full-screen prediction provides only one possible target location.
To explore alternative locations, we obtain one deterministic prediction
through greedy decoding and $n$ stochastic predictions sampled at
temperature $T$. We discard invalid sampled outputs and merge duplicate
coordinate predictions, retaining the sequence with the highest
log-probability. Let $N\leq n$ denote the number of retained samples and
$\lambda_j$ the sequence log-probability of sample $j$. We assign proposal
weights as
\begin{equation}
  w_0=1,\qquad
  w_j=\frac{\exp(\lambda_j)}{\sum_{r=1}^{N}\exp(\lambda_r)},
  \quad j=1,\ldots,N,
  \label{eq:weight}
\end{equation}
where $w_0$ applies to the deterministic prediction when it is valid. This
weighting preserves the model's deterministic prediction while assigning
the sampled alternatives an equal total weight.

Several predictions may refer to the same interface element with small
coordinate differences. To reduce redundant inspection, we cluster
predictions by spatial proximity, processing them in descending order of
weight. A prediction joins the first cluster whose seed lies within
$\delta$ merged visual tokens along both axes of the rendered view;
otherwise, it initializes a new cluster. Each cluster retains its seed as
the representative, and point-only predictions receive small centered boxes
for subsequent cropping.

For a cluster with member set $\mathcal I_h$, its hypothesis weight is
\begin{equation}
  M_h=\sum_{j\in\mathcal{I}_h}w_j.
  \label{eq:hyp_weight}
\end{equation}

Let $\mathcal J_1$ index the resulting clusters and $b_h$ denote the
representative box of cluster $h$. The first-level hypothesis set is
\begin{equation}
  \mathcal H_1=\bigl\{b_h\mid
    h\in\operatorname{TopK}_{r\in\mathcal J_1}(M_r,k)\bigr\},
  \label{eq:hyp_select}
\end{equation}
where $\operatorname{TopK}$ returns the indices of up to $k$ highest-weight
clusters. These weights determine which regions receive further inspection.
Final candidate selection uses the verification scores obtained after local
re-localization.

% ---------------------------------------------------------------------------
\subsection{Zoomed Localization and Verification}
\label{sec:zoom}
\label{sec:verify}

Local inspection provides finer visual evidence while preserving context
around a possible target. For each hypothesis $b_h$, we construct a crop
$c_h$ from the original screenshot. Let $W$ and $H$ denote the screenshot
dimensions, and $w_b$ and $h_b$ the hypothesis box dimensions in screenshot
pixels. The initial crop is centered on $b_h$, with width and height
\begin{equation}
  w_c=\max(\beta w_b,\rho W),\qquad
  h_c=\max(\beta h_b,\rho H),
  \label{eq:crop}
\end{equation}
where $\beta$ controls the surrounding context and $\rho$ specifies the
minimum crop extent. We expand the crop to match the screenshot's aspect
ratio and adjust it to remain within the image boundaries.

To control magnification, we set the crop's requested minimum pixel budget
using its adjusted dimensions $w_c$ and $h_c$:
\begin{equation}
  P(c_h)=(\kappa s_0)^2w_ch_c,
  \label{eq:budget}
\end{equation}
Here, $s_0=\min(1,\sqrt{P_{\max}/(WH)})$ is the reference scale for the
checkpoint's full-screen pixel budget $P_{\max}$, and $\kappa$ controls the
intended magnification. The processor clips the budget
and aligns the rendered dimensions with its visual-token grid. We query
$G$ on the resulting view $v_h$ using deterministic decoding. Let
$u_h=(u_h^x,u_h^y)$ be the predicted box center in normalized view
coordinates, and $(x_h^0,y_h^0)$ the crop's upper-left corner in screenshot
pixels. The candidate click in normalized screenshot coordinates is
\begin{equation}
  p_h=\left(\frac{x_h^0+u_h^x w_c}{W},\;
             \frac{y_h^0+u_h^y h_c}{H}\right).
  \label{eq:map}
\end{equation}

A magnified view can still contain several visually similar elements, so
the resulting candidate requires assessment against the instruction.
Following visual mark prompting~\cite{yang2023som}, we draw a red ring at
the predicted location to obtain a marked view $z_h$. The same model is
asked whether the marked element matches $q$, with a Yes/No response. Let
$\ell_{\mathrm{Yes}}(z_h,q)$ and $\ell_{\mathrm{No}}(z_h,q)$ denote the
estimated first-token log-probability scores, each taking the maximum over
the available single-token variants of its label. We define the
verification score as
\begin{equation}
  m_h=\ell_{\mathrm{Yes}}(z_h,q)-\ell_{\mathrm{No}}(z_h,q).
  \label{eq:margin}
\end{equation}

A larger score indicates stronger relative support for a match between the
marked element and the instruction. We use this score to rank candidates
within the same screenshot.

% ---------------------------------------------------------------------------
\subsection{Verification-Guided Refinement and Selection}
\label{sec:reopen}

A high-scoring first-level candidate may still be mislocalized among nearby
interface elements. \method therefore allocates one additional round of
exploration to the view associated with the highest-scoring candidate.
Let $\mathcal C_1$ index the valid first-level candidates, each with a click
$p_i$, verification score $m_i$, and local view $v_i$. We choose the view to
reopen by
\begin{equation}
  r=\operatorname*{arg\,max}_{i\in\mathcal C_1}m_i,
  \qquad v_{\mathrm{reopen}}=v_r.
  \label{eq:reopen}
\end{equation}

Within this view, we draw $n'$ stochastic predictions at temperature $T$,
filter and deduplicate them, and normalize their sequence-based weights.
We then apply spatial clustering using the visual-token spacing of this
local view and retain up to $k'$ hypotheses.

For each retained hypothesis, we construct a new view from the original
screenshot using a smaller crop floor $\rho'<\rho$. The context factor
$\beta$ and magnification rule remain unchanged. Each view undergoes the
localization and verification procedure in \S\ref{sec:zoom}, producing
additional candidate clicks. First-level candidates are retained so that a
position found earlier remains available even when the second round
produces different predictions.

Let $\mathcal C_2$ index the valid second-level candidates. The final
prediction is
\begin{equation}
  i^\star=\operatorname*{arg\,max}_{i\in\mathcal C_1\cup\mathcal C_2}m_i,
  \qquad p=p_{i^\star}.
  \label{eq:select}
\end{equation}

A second-level candidate replaces the best first-level candidate only when
it receives a higher verification score; ties retain the first-level
candidate. The final pool consists of re-localized candidates from both
levels, while the initial full-screen predictions serve as spatial
proposals.

The number of generated output sequences per screenshot is bounded by
\begin{equation}
  N_{\mathrm{seq}}\leq 1+n+2k+n'+2k',
  \label{eq:sequence_count}
\end{equation}
including at most $k+k'$ single-token verification responses. Fewer retained
hypotheses reduce the actual count. The inference cost also depends on the
rendered view sizes and output lengths.

% Main-text pseudocode, included at the end of Sec. 3.5.
Algorithm~\ref{alg:method} summarizes the procedure. \textsc{Loc} maps a
deterministically decoded box to normalized screenshot coordinates using
Eq.~\eqref{eq:map} on both corners. \textsc{Sample} returns valid,
deduplicated boxes in the same coordinates with weights from
Eq.~\eqref{eq:weight}; \textsc{Cluster} implements
Eqs.~\eqref{eq:hyp_weight}--\eqref{eq:hyp_select} using its input view's
visual-token spacing. \textsc{Mark} projects the stored box back into the
local view before adding the ring; \textsc{Verify} computes
Eq.~\eqref{eq:margin}.

\begin{algorithm}[!htbp]
  \caption{\method for one screenshot}
  \label{alg:method}
  \small
  \begin{algorithmic}[1]
    \Require screenshot $I$, instruction $q$, frozen model $G$
    \State $v_0\gets\textsc{FullScreen}(I)$
    \State $a_0\gets\textsc{Loc}(v_0,q)$;\quad
           $A\gets\textsc{Sample}(v_0,q,n,T)$
    \If{$a_0$ is valid}
      \State $A\gets A\cup\{(a_0,1)\}$
    \EndIf
    \State $\mathcal H_1\gets\textsc{Cluster}(A,v_0,\delta,k)$
    \State $\mathcal C_1\gets\textsc{LocalizeVerify}(\mathcal H_1,\rho)$
    \If{$\mathcal C_1=\emptyset$}
      \State \Return failure
    \EndIf
    \State Select the reopening index $r$ by Eq.~\eqref{eq:reopen}
    \State $A'\gets\textsc{Sample}(v_r,q,n',T)$ \Comment{re-open}
    \State $\mathcal H_2\gets\textsc{Cluster}(A',v_r,\delta,k')$
    \State $\mathcal C_2\gets\textsc{LocalizeVerify}(\mathcal H_2,\rho')$
    \State Select $i^\star$ by Eq.~\eqref{eq:select}, favoring $\mathcal C_1$ in ties
    \State \Return $p_{i^\star}$
    \Statex
    \Function{LocalizeVerify}{$\mathcal H,\alpha$}
      \State $\mathcal C\gets\emptyset$
      \For{each representative box $b\in\mathcal H$}
        \State $c\gets\textsc{Crop}(I,b,\beta,\alpha)$
        \State $v\gets\textsc{Render}(I,c,P(c))$
        \State $\hat b\gets\textsc{Loc}(v,q)$
        \If{$\hat b$ is valid}
          \State Assign a new candidate index $j$;\quad $v_j\gets v$
          \State $p_j\gets$ center of $\hat b$
          \State $m_j\gets\textsc{Verify}(\textsc{Mark}(v,\hat b),q)$
          \State $\mathcal C\gets\mathcal C\cup\{j\}$
        \EndIf
      \EndFor
      \State \Return $\mathcal C$
    \EndFunction
  \end{algorithmic}
\end{algorithm}

\section{Experiments}
\label{sec:experiments}
\begin{table*}[t]
  \caption{Grounding accuracy (\%) across eight benchmark settings.
  Rows with + \method apply our method to the model above.
  Bold marks the better result within each pair; superscripts show changes
  in percentage points. Mind2Web uses element-only instructions.}
  \label{tab:main}
  \centering
  % BEGIN GENERATED TABLE CONTENT
% Generated by tools/export_main_results_tables.py from the reviewed data snapshot.
\begingroup
\definecolor{webshade}{HTML}{EDF3FA}
\definecolor{methodblue}{HTML}{254D78}
\definecolor{gain}{HTML}{247565}
\definecolor{loss}{HTML}{9A4A47}
\definecolor{muted}{HTML}{545A62}
% Center the score itself; the gain is an annotation to its right.
\newcommand{\mainScore}[2]{\makebox[36pt][c]{#1\rlap{\textsuperscript{#2}}}}
\fontsize{8.5}{10.4}\selectfont
\setlength{\tabcolsep}{3pt}
\renewcommand{\arraystretch}{1.13}
\begin{tabularx}{\linewidth}{>{\raggedright\arraybackslash}X >{\columncolor{webshade}}c >{\columncolor{webshade}}c *{6}{c}}
\toprule
\multicolumn{1}{c}{} & \multicolumn{2}{c}{\cellcolor{webshade}\textcolor{methodblue}{\textbf{Web}}} & \multicolumn{6}{c}{\textbf{Desktop and cross-platform}} \\
\cmidrule(lr){2-3}\cmidrule(l){4-9}
\textbf{Model} & \textbf{Mind2Web} & \textbf{VisualWebBench} & \textbf{SS-Pro} & \textbf{OSW-G} & \textbf{OSW-GR} & \textbf{UI-V} & \textbf{WinDesk} & \textbf{MMB-GUI} \\
\midrule
\multicolumn{9}{l}{\textit{\textcolor{muted}{General-purpose MLLMs}}} \\[1pt]
Qwen2.5-VL-3B & \mainScore{65.9}{} & \mainScore{88.0}{} & \mainScore{25.1}{} & \mainScore{30.7}{} & \mainScore{36.7}{} & \mainScore{13.9}{} & \mainScore{48.4}{} & \mainScore{61.8}{} \\
\hspace{1em}\textcolor{methodblue}{+ MHVF} & \mainScore{\textbf{71.2}}{\textcolor{gain}{+5.3}} & \mainScore{\textbf{88.2}}{\textcolor{gain}{+0.2}} & \mainScore{\textbf{50.0}}{\textcolor{gain}{+24.9}} & \mainScore{\textbf{44.7}}{\textcolor{gain}{+14.0}} & \mainScore{\textbf{55.1}}{\textcolor{gain}{+18.4}} & \mainScore{\textbf{23.4}}{\textcolor{gain}{+9.5}} & \mainScore{\textbf{57.3}}{\textcolor{gain}{+8.9}} & \mainScore{\textbf{68.4}}{\textcolor{gain}{+6.6}} \\
\addlinespace[3pt]
Qwen2.5-VL-7B & \mainScore{69.0}{} & \mainScore{85.7}{} & \mainScore{23.6}{} & \mainScore{33.9}{} & \mainScore{38.7}{} & \mainScore{15.6}{} & \mainScore{50.3}{} & \mainScore{67.6}{} \\
\hspace{1em}\textcolor{methodblue}{+ MHVF} & \mainScore{\textbf{73.5}}{\textcolor{gain}{+4.5}} & \mainScore{\textbf{87.8}}{\textcolor{gain}{+2.1}} & \mainScore{\textbf{53.6}}{\textcolor{gain}{+30.0}} & \mainScore{\textbf{52.0}}{\textcolor{gain}{+18.1}} & \mainScore{\textbf{62.2}}{\textcolor{gain}{+23.6}} & \mainScore{\textbf{31.2}}{\textcolor{gain}{+15.6}} & \mainScore{\textbf{60.7}}{\textcolor{gain}{+10.4}} & \mainScore{\textbf{76.3}}{\textcolor{gain}{+8.7}} \\
\addlinespace[3pt]
\multicolumn{9}{l}{\textit{\textcolor{muted}{GUI-specific models}}} \\[1pt]
GTA1-7B & \mainScore{\textbf{76.1}}{} & \mainScore{\textbf{91.3}}{} & \mainScore{49.3}{} & \mainScore{55.1}{} & \mainScore{67.6}{} & \mainScore{26.1}{} & \mainScore{62.4}{} & \mainScore{\textbf{79.6}}{} \\
\hspace{1em}\textcolor{methodblue}{+ MHVF} & \mainScore{75.2}{\textcolor{loss}{-0.8}} & \mainScore{91.1}{\textcolor{loss}{-0.2}} & \mainScore{\textbf{59.1}}{\textcolor{gain}{+9.8}} & \mainScore{\textbf{58.0}}{\textcolor{gain}{+2.8}} & \mainScore{\textbf{68.1}}{\textcolor{gain}{+0.5}} & \mainScore{\textbf{34.0}}{\textcolor{gain}{+7.9}} & \mainScore{\textbf{63.2}}{\textcolor{gain}{+0.8}} & \mainScore{76.5}{\textcolor{loss}{-3.0}} \\
\addlinespace[3pt]
GUI-G2-7B & \mainScore{76.6}{} & \mainScore{90.5}{} & \mainScore{48.3}{} & \mainScore{52.0}{} & \mainScore{61.3}{} & \mainScore{25.9}{} & \mainScore{61.9}{} & \mainScore{79.5}{} \\
\hspace{1em}\textcolor{methodblue}{+ MHVF} & \mainScore{\textbf{79.2}}{\textcolor{gain}{+2.7}} & \mainScore{\textbf{92.1}}{\textcolor{gain}{+1.6}} & \mainScore{\textbf{64.1}}{\textcolor{gain}{+15.9}} & \mainScore{\textbf{60.3}}{\textcolor{gain}{+8.3}} & \mainScore{\textbf{69.1}}{\textcolor{gain}{+7.8}} & \mainScore{\textbf{38.3}}{\textcolor{gain}{+12.3}} & \mainScore{\textbf{64.0}}{\textcolor{gain}{+2.2}} & \mainScore{\textbf{82.3}}{\textcolor{gain}{+2.8}} \\
\addlinespace[3pt]
UI-Venus-Ground-7B & \mainScore{75.7}{} & \mainScore{90.7}{} & \mainScore{50.3}{} & \mainScore{54.1}{} & \mainScore{63.1}{} & \mainScore{26.7}{} & \mainScore{62.0}{} & \mainScore{80.2}{} \\
\hspace{1em}\textcolor{methodblue}{+ MHVF} & \mainScore{\textbf{80.0}}{\textcolor{gain}{+4.3}} & \mainScore{\textbf{92.8}}{\textcolor{gain}{+2.1}} & \mainScore{\textbf{66.6}}{\textcolor{gain}{+16.3}} & \mainScore{\textbf{59.8}}{\textcolor{gain}{+5.7}} & \mainScore{\textbf{70.7}}{\textcolor{gain}{+7.6}} & \mainScore{\textbf{38.8}}{\textcolor{gain}{+12.1}} & \mainScore{\textbf{64.3}}{\textcolor{gain}{+2.3}} & \mainScore{\textbf{82.8}}{\textcolor{gain}{+2.6}} \\
\bottomrule
\end{tabularx}
\endgroup
% END GENERATED TABLE CONTENT
\end{table*}

% Queue the results tables early so they remain near Section 4.2.
\begin{table}[!tb]
  \caption{Mind2Web grounding accuracy (\%) by test split.
  (a) Element descriptions only; (b) task instructions plus element descriptions.
  Bold marks the better result
  within each model pair. Overall aggregates all instances.}
  \label{tab:mind2web_splits}
  \centering
  % BEGIN GENERATED TABLE CONTENT
% Generated by tools/export_main_results_tables.py; single-column layout.
\begingroup
\definecolor{elementshade}{HTML}{F2F3F5}
\definecolor{taskshade}{HTML}{EDF3FA}
\definecolor{methodblue}{HTML}{254D78}
\definecolor{muted}{HTML}{606772}
\fontsize{8.2}{10}\selectfont
\setlength{\tabcolsep}{1.5pt}
\renewcommand{\arraystretch}{1.1}
\newcommand{\splitHead}[1]{{\fontsize{7.6}{9.5}\selectfont\bfseries #1}}
\begin{tabularx}{\linewidth}{@{}>{\raggedright\arraybackslash}Xcccc@{}}
\toprule
\splitHead{Model} & \splitHead{Cross-task} & \splitHead{Cross-website} & \splitHead{Cross-domain} & \splitHead{Overall} \\
\midrule
\multicolumn{5}{@{}>{\columncolor{elementshade}[0pt][0pt]}l@{}}{\textbf{(a) Element only}} \\[2pt]
Qwen2.5-VL-7B & 66.1 & 62.9 & 71.4 & 69.0 \\
\hspace{0.7em}\textcolor{methodblue}{+ MHVF} & \textbf{71.2} & \textbf{74.6} & \textbf{74.0} & \textbf{73.5} \\
\addlinespace[3pt]
UI-Venus-Ground-7B & 75.1 & 71.7 & 76.8 & 75.7 \\
\hspace{0.7em}\textcolor{methodblue}{+ MHVF} & \textbf{77.5} & \textbf{80.6} & \textbf{80.6} & \textbf{80.0} \\
\addlinespace[3pt]
\multicolumn{5}{@{}>{\columncolor{taskshade}[0pt][0pt]}l@{}}{\textbf{(b) Task + element}} \\[2pt]
Qwen2.5-VL-7B & 58.2 & 51.2 & 60.5 & 58.6 \\
\hspace{0.7em}\textcolor{methodblue}{+ MHVF} & \textbf{68.6} & \textbf{65.2} & \textbf{69.0} & \textbf{68.3} \\
\addlinespace[3pt]
UI-Venus-Ground-7B & 75.2 & 71.2 & 74.9 & 74.4 \\
\hspace{0.7em}\textcolor{methodblue}{+ MHVF} & \textbf{80.1} & \textbf{80.0} & \textbf{79.2} & \textbf{79.5} \\
\bottomrule
\end{tabularx}
\endgroup
% END GENERATED TABLE CONTENT
\end{table}

\begin{table*}[t]
  \caption{ScreenSpot-Pro accuracy (\%) of existing methods and MHVF, grouped by backbone. Shaded rows show original models and + MHVF; -- denotes an unreported configuration.}
  \label{tab:sota}
  \centering
  \definecolor{sotashade}{HTML}{EDF3FA}
  \definecolor{sotablue}{HTML}{254D78}
  \fontsize{8.5}{10.2}\selectfont
  \setlength{\tabcolsep}{3pt}
  \renewcommand{\arraystretch}{1.04}
  \begin{minipage}[t]{0.515\textwidth}
    \vspace{0pt}
    \textbf{(a) General-purpose backbone: Qwen2.5-VL}\par\vspace{5pt}
    \begin{tabularx}{\linewidth}{>{\raggedright\arraybackslash}X *{2}{>{\centering\arraybackslash}p{32pt}}}
    \toprule
\textbf{Method} & \textbf{3B} & \textbf{7B} \\
    \midrule
    \multicolumn{3}{l}{\emph{Forward-inference methods}} \\
    GUI-RC~\cite{du2026guirc} & 23.6 & 24.0 \\
    RegionFocus~\cite{luo2025regionfocus} & -- & 32.1 \\
    VPSG~\cite{tao2025vpsg} & 13.3 & 19.1 \\
    UI-Zoomer~\cite{tang2026uizoomer} & -- & 41.0 \\
    ZoomClick~\cite{jiang2025zoomclick} & -- & 44.0 \\
    Trifuse~\cite{ma2026trifuse} & 18.9 & 29.7 \\
    DRS-GUI~\cite{liu2026drsgui} & 28.7 & 40.9 \\
    AutoFocus~\cite{yao2026autofocus} & -- & 35.4 \\
    \midrule
    \multicolumn{3}{l}{\emph{Latent optimization}} \\
    ZoomUI~\cite{liu2026zoomui} & 40.2 & 52.8 \\
    \midrule
    \multicolumn{3}{l}{\emph{Parameter adaptation or additional training}} \\
    GUI-RCPO~\cite{du2026guirc} & 24.7 & 25.9 \\
    CAL~\cite{liu2026canl} & 32.1 & 32.7 \\
    CANL~\cite{liu2026canl} & 32.7 & 33.8 \\
    R-OPSD (MMBench-GUI)~\cite{xuan2026selfevolving} & 30.1 & 32.3 \\
    GUI-Spotlight~\cite{lei2025guispotlight} & -- & 38.7 \\
    ReGUIDE (v2)~\cite{lee2025reguide} & 44.3 & 44.4 \\
    \midrule
    \rowcolor{sotashade}
    Qwen2.5-VL & 25.1 & 23.6 \\
    \rowcolor{sotashade}
    \hspace{0.7em}\textcolor{sotablue}{\textbf{+ \method}} & \textbf{50.0} & \textbf{53.6} \\
    \bottomrule
    \end{tabularx}
  \end{minipage}\hfill
  \begin{minipage}[t]{0.46\textwidth}
    \vspace{0pt}
    \textbf{(b) GUI-specific backbones}\par\vspace{5pt}
    \begin{tabularx}{\linewidth}{>{\raggedright\arraybackslash}X >{\centering\arraybackslash}p{47pt}}
    \toprule
\textbf{Method} & \textbf{Acc. (\%)} $\uparrow$ \\
    \midrule
    \multicolumn{2}{l}{\emph{GTA1-7B}} \\
    MVP~\cite{zhang2025mvp} & 61.7 \\
    AutoFocus~\cite{yao2026autofocus} & 54.8 \\
    MEGA-GUI~\cite{kwak2025megagui} & 68.8 \\
    \addlinespace[2pt]
    \rowcolor{sotashade}
    GTA1-7B & 49.3 \\
    \rowcolor{sotashade}
    \hspace{0.7em}\textcolor{sotablue}{\textbf{+ \method}} & \textbf{59.1} \\
    \addlinespace[4pt]
    \midrule
    \multicolumn{2}{l}{\emph{GUI-G2-7B}} \\
    GUI-RC~\cite{du2026guirc} & 47.9 \\
    UI-Zoomer~\cite{tang2026uizoomer} & 61.4 \\
    \addlinespace[2pt]
    \rowcolor{sotashade}
    GUI-G2-7B & 48.3 \\
    \rowcolor{sotashade}
    \hspace{0.7em}\textcolor{sotablue}{\textbf{+ \method}} & \textbf{64.1} \\
    \addlinespace[4pt]
    \midrule
    \multicolumn{2}{l}{\emph{UI-Venus-Ground-7B}} \\
    UI-Zoomer~\cite{tang2026uizoomer} & 61.8 \\
    ZoomClick~\cite{jiang2025zoomclick} & 65.7 \\
    AutoFocus~\cite{yao2026autofocus} & 67.8 \\
    \addlinespace[2pt]
    \rowcolor{sotashade}
    UI-Venus-Ground-7B & 50.3 \\
    \rowcolor{sotashade}
    \hspace{0.7em}\textcolor{sotablue}{\textbf{+ \method}} & \textbf{66.6} \\
    \bottomrule
    \end{tabularx}
  \end{minipage}
\end{table*}

\subsection{Experimental Setup}
\label{sec:setup}

\begingroup
\setlength{\emergencystretch}{2em}
\paragraph{Metrics and Benchmarks.}
We evaluate \method using grounding accuracy: a prediction is correct if
its coordinate falls within the annotated target region. We report results
on eight benchmark settings spanning web, desktop, and cross-platform
interfaces: Mind2Web~\cite{deng2023mind2web,zheng2024seeact},
VisualWebBench~\cite{liu2024visualwebbench},
ScreenSpot-Pro~\cite{li2025sspro}, OSWorld-G (OSW-G) and
OSWorld-G-Refine (OSW-GR)~\cite{xie2025osworldg},
UI-Vision~\cite{nayak2025uivision}, WinDeskGround~\cite{windeskground},
and MMBench-GUI L2~\cite{wang2025mmbenchgui}.
On Mind2Web, we evaluate individual grounding steps across its three test
splits, using element descriptions alone or combined with task instructions.
For VisualWebBench, we use the element-grounding and action-grounding
subtasks with coordinate predictions on screenshots without option markers.
\par

\paragraph{Baseline Models.}
We evaluate \method on both general-purpose open-source MLLMs and
GUI-specific models. For general-purpose MLLMs, we adopt
Qwen2.5-VL-3B-Instruct and Qwen2.5-VL-7B-Instruct~\cite{bai2025qwen25vl}.
For GUI-specific models, we use GTA1-7B~\cite{yang2025gta1},
GUI-G2-7B~\cite{tang2025guig2}, and
UI-Venus-Ground-7B~\cite{gu2025uivenus}, which are trained for GUI grounding.
All five models are based on the Qwen2.5-VL architecture. These models
allow us to assess the robustness and effectiveness of \method as a
training-free enhancement to GUI grounding.
\par

\paragraph{Implementation Details.}
We run experiments with vLLM~\cite{kwon2023vllm} on NVIDIA A800 80GB GPUs.
We set the sampling temperature to 1.0, drawing eight samples and retaining
up to four hypotheses at the first level, and six samples and up to three
hypotheses at the second level. The target magnification is $\kappa=3$.
We apply the same search configuration across all models and benchmarks.
\par
\endgroup

\subsection{Main Results}
\label{sec:main}

\paragraph{Performance on Web Benchmarks.}
\method improves four of the five models on each web benchmark
(Table~\ref{tab:main}). On Mind2Web, the gains range from 2.7 to 5.3
percentage points for these models; UI-Venus-Ground-7B improves from
75.7\% to 80.0\%. The improvements extend across all three test splits
for Qwen2.5-VL-7B and UI-Venus-Ground-7B under both instruction formats
(Table~\ref{tab:mind2web_splits}). With task instructions, the largest gains
occur on the cross-website split, reaching 14.1 and 8.8 points,
respectively. VisualWebBench shows smaller gains, with UI-Venus-Ground-7B
improving from 90.7\% to 92.8\%. GTA1-7B is the exception on both web
benchmarks, indicating that additional search does not benefit every
grounding model equally.

\paragraph{Performance on Desktop and Cross-Platform Benchmarks.}
The improvements also hold on ScreenSpot-Pro, OSW-G, OSW-GR, UI-Vision,
and WinDeskGround, where all five models benefit (Table~\ref{tab:main}).
ScreenSpot-Pro shows large gains for both general-purpose and GUI-specific
models: Qwen2.5-VL-7B improves from 23.6\% to 53.6\%, while
UI-Venus-Ground-7B improves from 50.3\% to 66.6\%. The latter also improves
from 26.7\% to 38.8\% on UI-Vision. These results extend the web findings
to professional desktop interfaces and other GUI domains. Improvements
are smaller on WinDeskGround and MMBench-GUI L2 for GUI-specific models;
GTA1-7B decreases from 79.6\% to 76.5\% on MMBench-GUI L2.

\paragraph{Comparison with Existing Methods.}
Table~\ref{tab:sota} compares \method with existing methods on
ScreenSpot-Pro, grouped by backbone and adaptation strategy. With
Qwen2.5-VL-7B, \method achieves 53.6\%, compared with the reported
40.9\% of DRS-GUI, 44.0\% of ZoomClick, and 52.8\% of ZoomUI.
On GUI-G2-7B, it reaches 64.1\%, compared with 61.4\% for UI-Zoomer.
Other methods obtain stronger results in some settings: AutoFocus reports
67.8\% on UI-Venus-Ground-7B, compared with our 66.6\%, and MEGA-GUI
reports 68.8\% on GTA1-7B, compared with our 59.1\%.
These reported results use different model configurations and compute;
MVP and VPSG also use different averaging conventions. We therefore use
Table~\ref{tab:main} to establish gains over each original model and
Section~\ref{sec:analysis} to examine the accuracy--cost trade-off.

\subsection{Ablation Study}
\label{sec:ablation}

\begin{table}[!tb]
  \caption{Component ablations (grounding accuracy, \%). The full method
  is rerun for these comparisons.}
  \label{tab:ablation}
  \centering
  \definecolor{ablationshade}{HTML}{EDF3FA}
  \fontsize{8}{9.5}\selectfont
  \setlength{\tabcolsep}{2.25pt}
  \renewcommand{\arraystretch}{1.14}
  \begin{tabularx}{\linewidth}{>{\raggedright\arraybackslash}Xccc}
    \toprule
    & \textbf{Mind2Web} & \multicolumn{2}{c}{\textbf{ScreenSpot-Pro}} \\
    \cmidrule(lr){2-2}\cmidrule(l){3-4}
    \textbf{Variant} & \shortstack{\textbf{UI-Venus-}\\\textbf{Ground-7B}}
            & \shortstack{\textbf{UI-Venus-}\\\textbf{Ground-7B}}
            & \shortstack{\textbf{Qwen2.5-}\\\textbf{VL-7B}} \\
    \midrule
    \rowcolor{ablationshade}
    \textbf{\method} & \textbf{80.0} & \textbf{66.6} & \textbf{53.9} \\
    \midrule
    Single initial hypothesis & 79.2 & 64.1 & 47.5 \\
    w/o Magnification ($\kappa{=}1$) & 78.2 & 61.6 & 41.4 \\
    w/o Reopening & 79.7 & 65.5 & 50.9 \\
    w/o Verification and Reopening & 78.9 & 61.9 & 45.7 \\
    Full-screen verification & 79.5 & 65.0 & 52.8 \\
    \bottomrule
  \end{tabularx}
\end{table}

Table~\ref{tab:ablation} evaluates the components of \method on Mind2Web
and ScreenSpot-Pro. Both multiple hypotheses and magnification contribute
to accuracy. Using a single initial hypothesis reduces accuracy by 0.8
points on Mind2Web and by 2.5 and 6.4 points on ScreenSpot-Pro for
UI-Venus-Ground-7B and Qwen2.5-VL-7B, respectively. Removing magnification
($\kappa=1$) while retaining cropping produces larger drops on
ScreenSpot-Pro, at 5.0 and 12.5 points. This comparison shows that
increasing the visual scale of a selected region helps beyond cropping
alone, particularly on high-resolution interfaces.

Verification also helps when the search depth is held fixed. Comparing the
two variants without reopening, verification raises accuracy over ranking
by hypothesis weight from 78.9\% to 79.7\% on Mind2Web and from 61.9\%
to 65.5\% and 45.7\% to 50.9\% on ScreenSpot-Pro. Reopening adds a
further 0.3, 1.1, and 3.0 points, respectively, with a modest improvement
on Mind2Web.
Full-screen verification yields losses of 0.4--1.6 points relative to
zoomed-view verification, with no statistically significant difference
for Qwen2.5-VL-7B on ScreenSpot-Pro or UI-Venus-Ground-7B on Mind2Web.
Thus, verification provides a clearer benefit than the choice of its
input view, while the value of reopening varies across settings.

\subsection{Further Analysis}
\label{sec:analysis}

\begin{figure}[!tb]
  \centering
  \includegraphics[width=\linewidth]{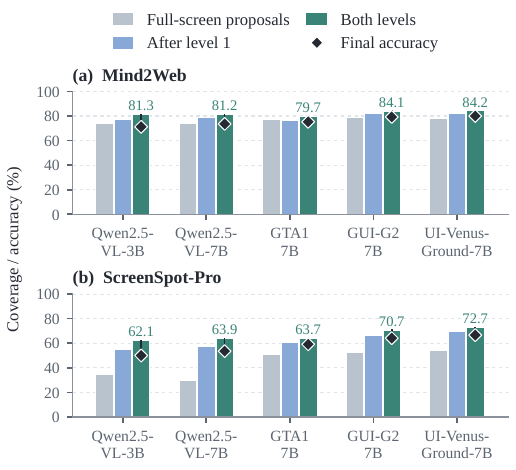}
  \caption{Candidate coverage and final accuracy. Bars show the fraction of samples with a correct candidate at each stage; diamonds mark final accuracy and segments show the remaining selection gap.}
  \label{fig:candidate_coverage}
  \Description{Grouped bars compare proposal coverage, coverage after level one, and coverage after both levels on Mind2Web and ScreenSpot-Pro. Diamonds mark final accuracy below the candidate-coverage upper bound.}
\end{figure}

\paragraph{Candidate Coverage and Final Selection.}
Local search finds correct candidates missed by the initial proposals,
but some are lost during final selection (Figure~\ref{fig:candidate_coverage}).
We measure coverage as the fraction of samples with at least one correct
candidate. On Mind2Web, UI-Venus-Ground-7B increases coverage from 78.3\%
for full-screen proposals to 84.2\% after both search levels, with 80.0\%
final accuracy. On ScreenSpot-Pro, first-level localization increases
coverage by 9.9--28.3 points across the five models, and reopening adds
another 2.8--7.5 points. The remaining gap between both-level coverage
and final accuracy is 4.2--12.1 points across the two benchmarks. Improving
selection therefore remains useful even after search has found the target.

\begin{figure}[!t]
  \centering
  \includegraphics[width=\linewidth]{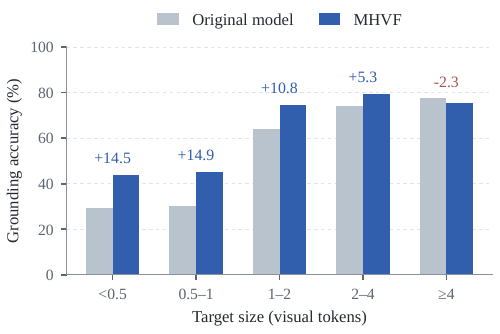}
  \caption{Accuracy by target size; labels show gains in percentage points. Results pool 39,445 sample--model pairs across three benchmarks and five backbones.}
  \label{fig:target_size}
  \Description{Grouped bars compare original-model and MHVF accuracy across five target-size bins. Gains are largest for targets smaller than one visual token, while the largest-target bin has a small decrease.}
\end{figure}

\begin{figure}[!t]
  \centering
  \includegraphics[width=\linewidth]{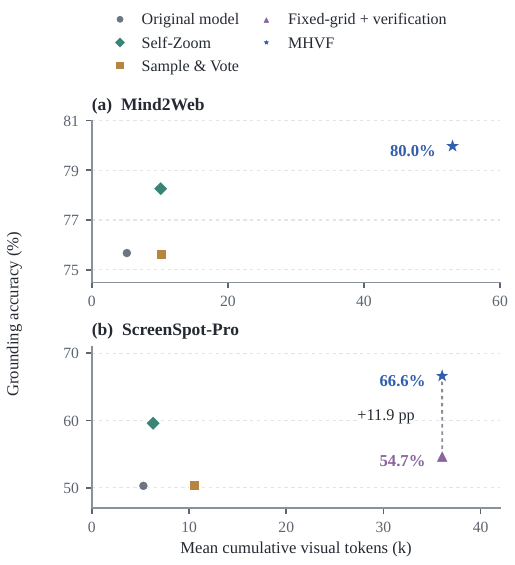}
  \caption{Accuracy versus visual input cost with UI-Venus-Ground-7B. Cost is the mean cumulative visual token estimate per sample, counting each input request once. The grid control and MHVF use comparable mean visual input on ScreenSpot-Pro.}
  \label{fig:accuracy_compute}
  \Description{Two scatter plots compare the original UI-Venus-Ground-7B,
  Self-Zoom, Sample and Vote, and MHVF, with an additional fixed-grid
  crop and verification control on ScreenSpot-Pro. Mind2Web is above
  ScreenSpot-Pro. At approximately 36,000 mean cumulative visual tokens,
  the grid control reaches 54.7 percent and MHVF reaches 66.6 percent;
  a dotted segment marks their 11.9 percentage point accuracy gap.}
\end{figure}

\paragraph{Target Size.}
The gains are concentrated on smaller targets (Figure~\ref{fig:target_size}).
We measure target size as the geometric mean of its width and height in
visual tokens under the reference full-screen scale, pooling records from
Mind2Web, VisualWebBench, and ScreenSpot-Pro across five models. Targets
smaller than one token gain 14.5--14.9 points, whereas targets of at least
four tokens show a net decrease of 2.3 points. This pattern is consistent
with the benefit of magnification for small elements, although the pooled
groups also differ in benchmark and model composition.

\paragraph{Accuracy and Visual Input Cost.}
\label{sec:control}
Figure~\ref{fig:accuracy_compute} compares the original UI-Venus-Ground-7B,
Self-Zoom with one additional zoomed grounding step, Sample-and-vote over
full-screen predictions, and \method. Using the full-method rerun from
Table~\ref{tab:ablation}, \method achieves 80.0\% on Mind2Web and 66.6\%
on ScreenSpot-Pro, exceeding Self-Zoom by 1.7 and 7.0 points.
Sample-and-vote remains close to the original model. The gains require
more visual input: relative to Self-Zoom, the mean cumulative visual token
estimate increases from 10.1k to 53.0k on Mind2Web and from 6.3k to 36.0k
on ScreenSpot-Pro.

On ScreenSpot-Pro, we also evaluate a \emph{fixed-grid crop and
verification} control. Fixed-grid cropping is used in ZoomClick's pre-zoom
stage~\cite{jiang2025zoomclick}; our control uses nine half-width,
half-height crops arranged on a $3\times3$ grid with 50\% overlap between
adjacent windows. It uses the same localizer and verifier as \method and
selects the highest-scoring candidate. At matched dataset-mean visual
input costs of approximately 36.0k tokens, the control achieves 54.7\%,
compared with 66.6\% for \method. This 11.9-point advantage supports
hypothesis-guided search over the fixed spatial coverage of this control.

\begin{figure}[!htbp]
  \centering
  \includegraphics[width=\linewidth]{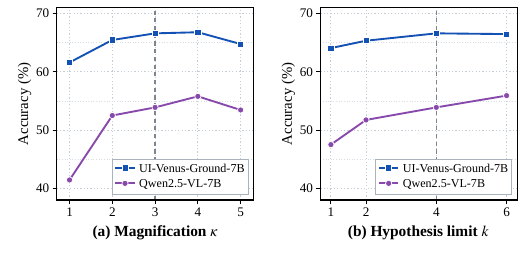}
  \caption{Parameter sensitivity on ScreenSpot-Pro: (a) magnification $\kappa$ and (b) the maximum number of first-level hypotheses $k$. Vertical dashed lines mark the adopted settings, $\kappa=3$ and $k=4$.}
  \label{fig:parameter_sensitivity}
  \Description{Two line plots show ScreenSpot-Pro accuracy versus magnification and the maximum number of first-level hypotheses for UI-Venus-Ground-7B and Qwen2.5-VL-7B.}
\end{figure}

\paragraph{Parameter Sensitivity.}
\label{sec:sensitivity}
Figure~\ref{fig:parameter_sensitivity} varies magnification $\kappa$ and
the first-level hypothesis limit $k$ on ScreenSpot-Pro. For both models,
accuracy increases up to $\kappa=4$ and then decreases at $\kappa=5$.
Increasing $\kappa$ from 3 to 4 adds 0.2 points for UI-Venus-Ground-7B
and 1.9 points for Qwen2.5-VL-7B, while increasing mean visual input by
approximately 45\% and 40\%, respectively. Increasing $k$ from 1 to 4
helps both models; at $k=6$, UI-Venus-Ground-7B remains nearly unchanged,
whereas Qwen2.5-VL-7B gains another 2.0 points. Thus, larger settings
increase cost without consistently improving accuracy. The adopted
$\kappa=3$ and $k=4$ provide a common configuration, while the preferred
accuracy--cost trade-off can differ by backbone.

\begin{figure}[!t]
  \centering
  \includegraphics[width=\linewidth]{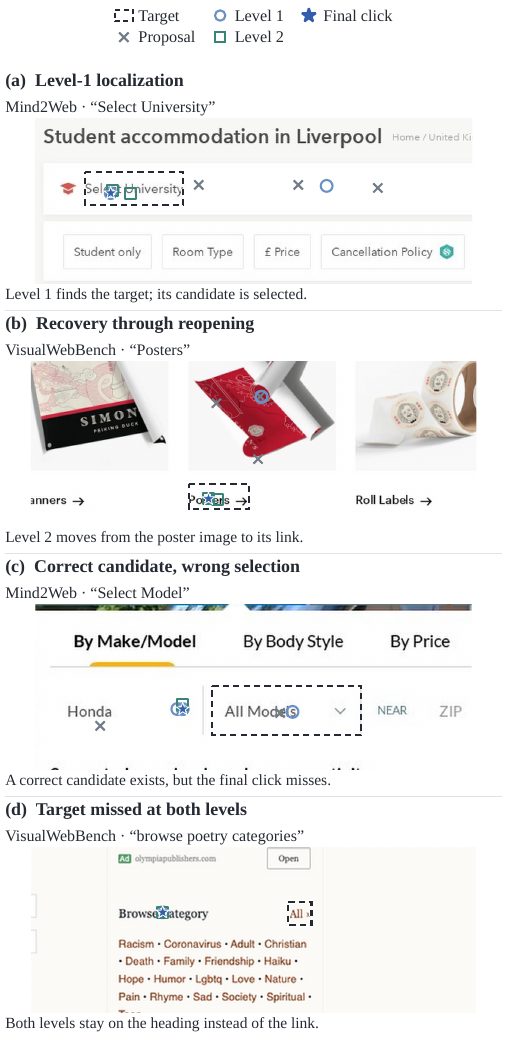}
  \caption{Web grounding cases with UI-Venus-Ground-7B on Mind2Web (a, c) and VisualWebBench (b, d): successful localization, recovery through reopening, selection failure, and missing candidate coverage. Markers show logged predictions from both levels; dashed boxes mark targets. Screenshots are cropped for display.}
  \label{fig:web_cases}
  \Description{Four webpage crops show level-1 localization for Select University, recovery of the Posters link at level 2, an incorrect final selection despite a correct candidate for Select Model, and both search levels missing the target link for browse poetry categories.}
\end{figure}

\paragraph{Web Grounding Cases.}
Figure~\ref{fig:web_cases} illustrates four outcomes on Mind2Web and
VisualWebBench. For ``Select University,'' level-1
localization finds the target missed by the full-screen proposals. For
``Posters,'' reopening reaches the text link after level 1 points to the
product image. The two failure cases expose different limitations:
``Select Model'' has a correct candidate that is not selected, whereas
``browse poetry categories'' remains on a heading instead of the target
link at both levels.

\section{Conclusion}
\label{sec:conclusion}

We presented \method, a training-free framework that combines multiple
spatial hypotheses with local visual verification for GUI grounding.
Using a single pretrained VLM, \method explores potential target regions,
re-localizes candidates in magnified views, and uses verification to guide
local refinement and final selection. Experiments with five models across
eight benchmark settings demonstrate improved accuracy in most settings,
covering web, desktop, and cross-platform interfaces. Candidate-level
analyses show that local search expands target coverage, while selecting
the correct candidate remains a source of error. This gap, together with
the additional visual input cost, motivates further work on more reliable
candidate verification and adaptive allocation of the search budget.

\clearpage
\bibliographystyle{ACM-Reference-Format}
\bibliography{refs}

\appendix
% The algorithm remains in the main method section.
\section{Implementation Details and View Geometry}
\label{app:view_geometry}

\paragraph{Search settings.}
Spatial clustering uses $\delta=2$ merged visual tokens. We set $\beta=3$
at both levels, with crop floors $\rho=0.25$ and $\rho'=0.10$, and reopen
one region. GTA1-7B point predictions receive centered boxes spanning
1.5\% of the input view along each axis, clipped before mapping to screenshot
coordinates. Localization uses greedy decoding with at most 32 output tokens.

\paragraph{Verification implementation.}
The point-verification prompt is: ``The red circle marks one UI element.
Is the circled element the one described as `\emph{instruction}'? Answer
Yes or No.'' We request one output token and the top-20 log-probabilities.
Each label score is the maximum over its available single-token variants,
including capitalization and leading spaces; absent labels receive $-30$.
These estimates are used in Eq.~\eqref{eq:margin} to rank candidates
without an acceptance threshold. The ring size follows the predicted box
extent and is bounded relative to the view dimensions.

\paragraph{Coordinate mapping.}
A view $v=(c,P)$ uses crop $c$ and minimum pixel budget $P$.
The bounds $c=(x_0,y_0,x_1,y_1)$ are in screenshot pixels.
The processor clips this budget to its
supported range, raises the maximum budget when needed, and aligns input
dimensions with its merged visual-token size $\tau$. Equation~\eqref{eq:map}
maps view-relative coordinates to normalized screenshot coordinates, using
$w_c=x_1-x_0$ and $h_c=y_1-y_0$. For boxes, the transformation applies to
both corners; point predictions use the extent specified above.

Let $P_{\max}$ denote the checkpoint's declared full-screen pixel budget.
For a screenshot of size $W\times H$, we define the reference scale
\begin{equation}
  s_0 = \min\Bigl(1,\sqrt{P_{\max}/(WH)}\Bigr).
  \label{eq:scale}
\end{equation}
This budget-based scale does not include processor minimum-size constraints
or dimension rounding. It controls the requested magnification. Spatial
clustering uses the visual-token spacing of the actual rendered view.

\paragraph{Method illustration.}
The example in Figure~\ref{fig:method} uses UI-Venus-Ground-7B on
ScreenSpot-Pro with the instruction ``add new tab.''
The display crops and red marks are reconstructed illustrations; the exact
second-level inference crop was not logged. The dashed target annotation
is shown for evaluation only and is not supplied to the grounder.

% Let the comparison follow naturally as bibliography length changes.
\section{Additional Literature Comparisons}
\label{app:comparison}

Table~\ref{tab:other_configurations} extends the ScreenSpot-Pro comparison
to additional backbones and multi-model systems, including ScreenSeekeR,
GMS, Re-Prefill, and GUI-Lens, as well as trained zooming methods.
Each row retains the configuration and accuracy reported by its source.

\begin{table}[H]
  \caption{Literature-reported ScreenSpot-Pro results (\%) for additional backbones and multi-model systems. Each row gives one reported configuration.}
  \label{tab:other_configurations}
  \centering
  \fontsize{8.5}{10.2}\selectfont
  \renewcommand{\arraystretch}{1.02}
  \setlength{\tabcolsep}{3pt}
  \renewcommand{\tabularxcolumn}[1]{m{#1}}
  \begin{tabularx}{\linewidth}{@{}>{\raggedright\arraybackslash}m{0.29\linewidth}>{\raggedright\arraybackslash}X>{\centering\arraybackslash}m{45pt}@{}}
\toprule
\textbf{Method} & \textbf{Model configuration} & \textbf{Acc. (\%)} $\uparrow$ \\
\midrule
\multicolumn{3}{@{}l}{\emph{Forward-inference methods}} \\
Iterative Focusing~\cite{li2025sspro} & OS-Atlas-7B & 31.0 \\
Iterative Narrowing~\cite{li2025sspro} & OS-Atlas-7B & 31.9 \\
ReGround~\cite{li2025sspro} & OS-Atlas-7B & 40.2 \\
ScreenSeekeR~\cite{li2025sspro} & OS-Atlas-7B + GPT-4o planner & 48.1 \\
ECP~\cite{lee2025ecp} & OS-Atlas-7B & 40.4 \\
DiMo-GUI~\cite{wu2025dimogui} & OS-Atlas-7B & 49.7 \\
GMS~\cite{li2025gms} & UGround-V1-7B + Gemini 2.5 Flash-Lite & 54.6 \\
BAMI~\cite{zhang2026bami} & TianXi-Action-7B + GPT-5 correction & 57.8 \\
Chain-of-Ground~\cite{li2025chainofground} & UI-TARS-1.5-7B $\to$ Qwen3-VL-235B $\to$ Qwen3-VL-32B & 68.4 \\
Re-Prefill~\cite{lin2026reprefill} & GUI-Owl-1.5-8B + zoom-in & 80.1 \\
GUI-Lens~\cite{fu2026guilens} & GPT-5.5 + coarse-to-fine cropping & 87.9 \\
DART~\cite{dart2026project} & KV-Ground-8B (project report) & 80.9 \\
\midrule
\multicolumn{3}{@{}l}{\emph{Parameter adaptation or additional training}} \\
AdaZoom-GUI~\cite{pei2026adazoom} & Trained 4B model + Qwen3.5-397B-A17B refiner & 76.8 \\
InnerZoom~\cite{liu2026innerzoom} & Qwen3-VL-4B + SFT/RL & 66.2 \\
\bottomrule
\end{tabularx}
\end{table}

% Keep the case-study heading, explanation, and wide figure on one page.
\clearpage
\twocolumn[{
  \begin{minipage}{\textwidth}
    \section{Extended Qualitative Cases}
\label{app:cases}

Figure~\ref{fig:qualitative_cases} presents six additional cases across
Mind2Web, VisualWebBench, and ScreenSpot-Pro. Panels (a) and (b) illustrate
localization among similar webpage elements and retention of a correct
level-1 candidate when reopening misses. Panels (c) and (d) show localization
and reopening in professional applications. Panels (e) and (f) contrast
selecting an incorrect candidate with failing to find a correct one.

% A non-floating figure inside the full-width appendix case-study block.
\par\medskip
\begin{minipage}{\linewidth}
  \captionsetup{type=figure}
  \centering
  \includegraphics[width=\linewidth]{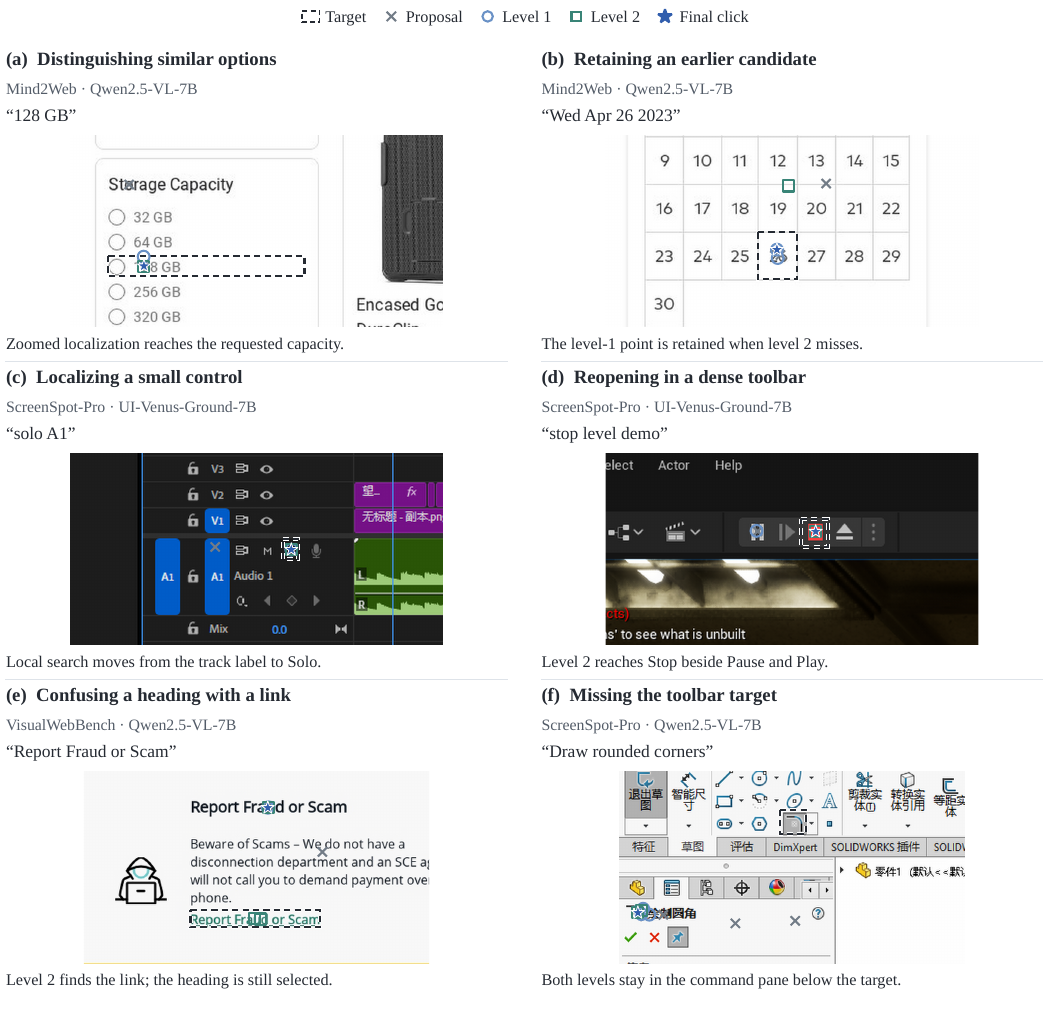}
  \caption{Additional cases with Qwen2.5-VL-7B and UI-Venus-Ground-7B: (a) localization among similar options, (b) retention of a level-1 candidate, (c) small-control localization, (d) recovery through reopening, (e) selection failure, and (f) missing candidate coverage. Markers show logged predictions from both levels; dashed boxes mark targets. Screenshots are cropped for display.}
  \label{fig:qualitative_cases}
  \Description{Six distinct screenshot crops show successful localization of the 128 GB option, retention of the correct April 26 calendar cell, localization of Premiere's Solo control, reopening to find Unreal Engine's Stop control, selection of a heading despite correct link candidates, and both levels missing a SolidWorks toolbar control.}
\end{minipage}

  \end{minipage}
}]

\end{document}